\documentclass[10pt,twocolumn]{article}

\usepackage[T1]{fontenc}
\usepackage{lmodern}
\usepackage[margin=0.72in,columnsep=0.25in]{geometry}
\usepackage{microtype}
\usepackage{graphicx}
\usepackage{booktabs}
\usepackage{multirow}
\usepackage{array}
\usepackage{amsmath,amssymb}
\usepackage{enumitem}
\usepackage{xcolor}
\usepackage{xurl}
\usepackage[hidelinks]{hyperref}
\usepackage[numbers,sort&compress]{natbib}
\usepackage{balance}

\definecolor{phoenixblue}{HTML}{2D6CDF}
\definecolor{phoenixnavy}{HTML}{18324A}
\definecolor{phoenixgray}{HTML}{596577}
\hypersetup{colorlinks=true,linkcolor=phoenixblue,citecolor=phoenixblue,urlcolor=phoenixblue}
\setlist[itemize]{leftmargin=*,nosep}
\setlist[enumerate]{leftmargin=*,nosep}
\newcommand{\system}{\textsc{Athar}}
\newcommand{\model}{\textsc{Phoenix}}
\newcommand{\cer}{\mathrm{CER}}
\newcommand{\wer}{\mathrm{WER}}
\newcommand{\pp}{\,\text{pp}}

\title{\vspace{-0.55in}\textbf{Phoenix and Athar: Compact Multi-Domain HTR with Evidence-Aware Review for Arabic Manuscripts}}
\author{Abdullah Ahmed Ali \quad Mohammed Thamer Abdulhadi \quad Ali Haider Safaa \quad Dhulfiqar Mahdi Wadi\\
\small University of Technology, Baghdad, Iraq\\
\small \href{https://huggingface.co/factlogic/phoenix-arabic-manuscript-htr}{huggingface.co/factlogic/phoenix-arabic-manuscript-htr}}
\date{August 2026}

\begin{document}
\maketitle

\begin{abstract}
Historical Arabic manuscript transcription is not only a recognition problem. A usable scholarly system must cope with shifting hands and layouts, preserve uncertain readings, distinguish visual evidence from linguistic plausibility, retrieve possible source parallels without inventing unique attribution, and record the researcher's final decision. We present \model{}, a 4.99-million-parameter CNN--BiLSTM--CTC recognizer, and \system{}, an evidence-aware review workflow built around it. \model{} is adapted across archival, Maghrebi, and historical manuscript domains using document-aware replay, an expanded 81-symbol codec, and forgetting guards that reject checkpoints that improve a new domain at unacceptable cost to previous domains. A historical specialist-model ablation shows that domain specialization yielded only marginal gains in development families (0.44\pp{} CER on Hard, 0.07\pp{} on Easy) and did not justify added routing complexity over a single unified recognizer. In a pre-specified held-out comparison against the preceding checkpoint, frozen before evaluation and scored with greedy decoding and raw references, \model{} reduced character error rate (CER) from 22.12\% to 17.86\% on 10,594 Agapet lines and from 17.72\% to 11.84\% on 11,684 Omar lines, while regressing from 10.39\% to 10.72\% on 164 TariMa lines. Across the two large held-out sets, character-weighted CER fell from 19.98\% to 14.93\%, a 25.3\% relative error reduction. A separate same-protocol development diagnostic found the lowest CER for \model{} on four of four comparable domains (9.59\% unweighted macro CER). Crucially, an $N$-best diagnostic on a balanced five-domain candidate pool reveals substantial remaining headroom (an oracle gap of 2.15 CER points; Beam 7.97\% vs.\ Oracle@25 5.82\%), while learned neural text rerankers (7.88\%), consensus MBR (8.01\%), CTC-posterior quality estimation (7.97\%), and local pre-CTC hidden-state quality estimation (7.97\%) recover less than 4\% of this gap. Because automated sequence selection fails to reliably bridge this oracle headroom, \system{} preserves the visual reading, exposes bounded alternatives, uses domain-conditioned local language models conservatively, retrieves parallels with explicit unique/ambiguous/abstain states, and exports auditable TEI and PAGE-XML records. The work demonstrates that manuscript HTR should be evaluated and deployed as auditable evidence management rather than silent text replacement.
\end{abstract}


\section{Introduction}
Digitizing a manuscript page as an image does not make its text searchable, comparable, or citable. Historical Arabic handwriting compounds the general difficulties of handwritten text recognition (HTR): connected letterforms, dot ambiguity, writer-specific glyphs, non-standard spelling, degraded supports, marginalia, frames, and reading orders that do not follow a single rectangular text block. Public resources such as Muharaf~\cite{saeed2024muharaf}, RASAM~\cite{vidalgorene2024rasam2}, TariMa~\cite{perrier2022tarima}, Agapet~\cite{ibrahim2025agapet}, and the Omar Al-Saleh shared-task data~\cite{hamoud2026nakba} have made multi-domain evaluation possible, but they also expose a central problem: a recognizer fine-tuned sequentially on one collection can lose accuracy on earlier collections.

For scholarly use, low average CER is necessary but insufficient. A single linguistically plausible correction can erase a meaningful scribal variant; a correct source parallel can still be an incorrect attribution when the phrase occurs in many places; and a confidence score can be mistaken for a calibrated probability. Existing open tools such as Kraken~\cite{kiessling2019kraken} and eScriptorium~\cite{stokes2021escriptorium} provide strong foundations for segmentation, HTR, correction, and export. 

Our investigation addresses two central research questions: (1) \emph{Can a compact recognizer be adapted across heterogeneous Arabic manuscript collections while preserving earlier-domain robustness without maintaining separate domain specialists?} and (2) \emph{Where does the remaining decoding error originate, and can modern automated sequence reranking or quality estimation reliably bridge the gap between candidate generation and candidate selection?}

This paper reports the current \model{} checkpoint (internal identifier \texttt{exp9}) and the current \system{} architecture. The main contributions are:

\begin{itemize}
  \item \textbf{Compact multi-domain recognition and specialization ablation.} A 4,988,946-parameter, 19.94-MB CNN--BiLSTM--CTC model with an 81-symbol output codec covers archival, Maghrebi, and historical Arabic handwriting. A historical ablation demonstrates that domain-specialist models yield only marginal gains ($0.07\text{--}0.44\pp{}$) over a single unified model.
  \item \textbf{Replay with explicit forgetting guards.} Training mixes new-domain data with prior-domain replay; checkpoints are selected by multi-domain guard performance under a pre-specified budget rather than newest-domain score alone.
  \item \textbf{Sealed, document-aware evaluation.} The release decision was made on 22,442 lines after freezing model identities and protocols. Results are strictly separated from development diagnostics and server benchmarks.
  \item \textbf{Empirical discovery of the candidate-selection bottleneck.} We demonstrate a substantial $N$-best oracle gap ($2.15\pp{}$ CER on development lines), while showing that learned neural text rerankers, minimum Bayes risk (MBR), CTC posterior quality estimation (QE), and local hidden-state QE recover very little of this headroom.
  \item \textbf{Evidence-aware review workflow.} Motivated by these candidate-selection limits, \system{} preserves the visual greedy reading, exposes bounded alternatives, uses domain-conditioned local language models conservatively, retrieves source parallels with unique/ambiguous/abstain states, prioritizes review, and exports auditable PAGE-XML, TEI, and training bundles.
\end{itemize}

\section{Related Work}
\subsection{Arabic manuscript HTR}
Muharaf provides more than 1,600 expert-transcribed historical page images and documents the diversity of archival Arabic hands~\cite{saeed2024muharaf}. RASAM and RASAM~2 focus on under-resourced Maghrebi scripts and heterogeneous manuscript production~\cite{vidalgorene2024rasam2}; TariMa provides diverse Maghrebi hands, layouts, and image qualities~\cite{perrier2022tarima}. Agapet contributes expert PAGE-XML for Christian Arabic manuscripts from the 13th to 17th centuries~\cite{ibrahim2025agapet}. The Omar Al-Saleh collection adds modern archival memoir handwriting and expert-verified line transcriptions~\cite{hamoud2026nakba}. These resources differ in period, hand, layout, transcription policy, and official split construction. Consequently, scores from different papers or splits are not automatically comparable.

Kraken is a script-agnostic OCR/HTR engine designed for humanities documents~\cite{kiessling2019kraken}. eScriptorium integrates Kraken into an open interface for segmentation, transcription, correction, training, and structured export~\cite{stokes2021escriptorium}. \system{} uses Kraken-compatible models and PAGE geometry, but its research focus is the controlled interaction among recognition, alternative readings, retrieval, and human evidence.

\subsection{Sequence recognition, continual learning, and quality estimation}
The recognizer follows the classical convolutional-recurrent sequence architecture. Bidirectional LSTMs model context in both directions~\cite{hochreiter1997lstm}; Connectionist Temporal Classification (CTC) marginalizes latent alignments between image frames and target symbols and removes the need for character-level segmentation~\cite{graves2006ctc}. This is particularly suitable for connected Arabic writing, where isolated character crops are neither natural nor stable.

Sequential domain fine-tuning risks catastrophic forgetting. Experience replay is a direct mechanism for preserving prior tasks by interleaving earlier examples with new-domain data~\cite{rolnick2019replay}. Our setting is not a formal online continual-learning benchmark: all permitted replay data are available offline. Nevertheless, the stability--plasticity problem is the same, and multi-domain guards make that trade-off measurable. Quality estimation and candidate reranking examine whether auxiliary neural models, minimum Bayes risk (MBR) consensus~\cite{kumar2004mbr}, or internal feature representations~\cite{specia2018qe} can re-score beam hypotheses without retraining the base acoustic/visual model.

\subsection{Abstention and scholarly evidence}
Selective classification formalizes the trade-off between coverage and risk by allowing a model to abstain~\cite{geifman2017selective}. \system{} applies the same principle to source attribution: no match, an ambiguous set of positions, and a unique attribution are distinct outcomes. PAGE-XML~\cite{pletschacher2010page} and TEI provide interoperable representations for geometry and text. OpenITI~\cite{nigst2025openiti} and IIIF~\cite{iiif2020presentation} motivate an extensible source and image layer, while provenance and licensing remain visible.

\section{System Overview}
\begin{figure*}[t]
  \centering
  \includegraphics[width=\textwidth]{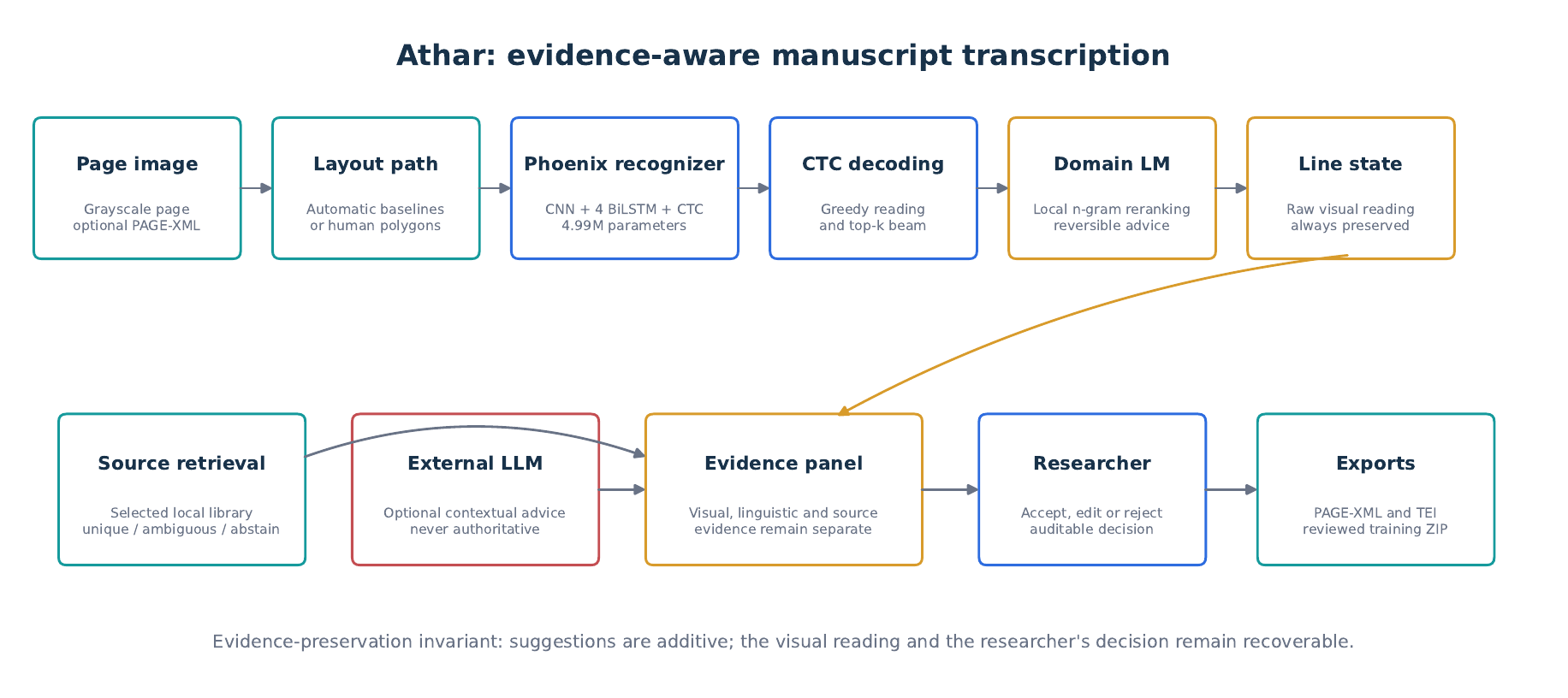}
  \caption{The \system{} architecture. Evidence channels are additive and labelled; they do not overwrite the visual reading or each other.}
  \label{fig:architecture}
\end{figure*}

Figure~\ref{fig:architecture} shows the current pipeline. A page follows either automatic baseline segmentation or a human/PAGE-XML path. Line images are recognized by \model{}. Greedy CTC decoding produces the preserved visual reading; beam decoding produces alternatives. A small domain-conditioned character $n$-gram model can rerank the same CTC candidates, but its scores remain relative and are not presented as calibrated correctness probabilities. Retrieval searches a project-selected local library. An optional external LLM can propose a contextual reading, explicitly labelled advisory. The evidence panel presents these channels separately, and the researcher accepts, edits, or rejects a reading before structured export.

The key design invariant is reversibility. The system stores \texttt{greedy\_text} even when a language-ranked suggestion is accepted automatically in a conservative band. Researcher decisions are keyed by page identity and line geometry, and model, image, polygon, decoder, and corpus hashes are incorporated into caches or manifests to prevent stale evidence from being attached to a different line.

\section{Phoenix Recognizer}
\subsection{Architecture}
\model{} accepts grayscale line images normalized to height 120. Four convolutional blocks extract visual features. Four bidirectional LSTM layers, with 200 units in each direction, convert the feature sequence into contextual representations. A linear CTC head predicts an 81-symbol codec plus the CTC blank. The complete model contains 4,988,946 trainable parameters and occupies 19,936,101 bytes in Kraken's \texttt{mlmodel} format. Its SHA-256 identity is \texttt{2896fef9d9665cbb...}~\cite{phoenix2026model}.

For an input line $x$ and output sequence $y$, CTC sums the probabilities of all frame-level paths $\pi$ that collapse to $y$:
\begin{equation}
P(y\mid x)=\sum_{\pi\in \mathcal{B}^{-1}(y)}\prod_{t=1}^{T}P(\pi_t\mid x),
\end{equation}
where $\mathcal{B}$ removes blanks and repeated labels. Greedy decoding takes the locally most likely frame labels. Beam decoding retains multiple paths and supplies bounded alternatives to the review layer.

\subsection{Multi-domain replay, specialist ablation, and codec expansion}
\label{sec:replay}
The earlier recognizer had a 65-symbol codec that covered core Arabic letters but omitted many digits and punctuation marks present in archival documents. \model{} expands the codec to 81 symbols based on observed training evidence rather than adding a generic multilingual alphabet. The goal is to regain relevant archival symbols without spending output capacity on unsupported scripts.

Training used document-aware data from five sources: Muharaf archival handwriting, RASAM Maghrebi manuscripts, TariMa Maghrebi/historical manuscripts, Agapet SA-418, and a calibration/training subset of Omar. Earlier-domain samples were replayed while learning new domains. 

A central architecture question was whether to train multiple domain-specialist models and route between them, or train a single unified multi-domain recognizer. A historical development ablation examined this trade-off:
\begin{itemize}
  \item \textbf{Hard manuscript family:} General recognizer achieved 14.14\% CER / 47.71\% WER; specialist model achieved 13.70\% CER / 45.61\% WER (an improvement of $0.44\pp{}$ CER).
  \item \textbf{Easy manuscript family:} General recognizer achieved 6.01\% CER / 22.69\% WER; specialist model achieved 5.94\% CER / 22.38\% WER (an improvement of $0.07\pp{}$ CER).
\end{itemize}
Under these historical development families, specialization produced only marginal gains ($0.07\text{--}0.44\pp{}$) and did not justify the substantial routing and multi-model management complexity for this project. This empirical finding motivated maintaining a single, compact unified recognizer.

The model-selection objective was a constrained choice across four primary development guards: Agapet, Omar, RASAM, and Muharaf. The decision rule was written and committed to version control before the candidate checkpoints' guard scores were computed. Relative to the incumbent checkpoint it required a mean gain of at least 0.1 percentage points across the four guards together with a regression of no more than 0.1 points on any single guard; a separate ceiling allowed no prior domain to regress by more than 0.5 points relative to the previously released checkpoint. A candidate with a lower newest-domain score was rejected when its degradation on prior domains exceeded this budget. A fifth TariMa development guard was measured but deliberately excluded from the rule, because the manuscripts it holds had been used to train the earlier checkpoint, so it compares memorization against generalization rather than two comparable models (Section~\ref{sec:dev}). The released checkpoint cleared the budget with a mean gain of 0.41 points and no regression on any of the four guards.

\paragraph{Post-release exploratory runs.} Two further training rounds were carried out after the released checkpoint was frozen. Both ran on external infrastructure; their executable checkpoints and complete producer and evaluation artifacts were not retained locally, so the figures below come from the project's summary records and could not be re-derived for this paper. No claim in this paper depends on them. A full extra epoch improved the four primary development guards by an average of 0.10 percentage points and slightly worsened the TariMa guard; it was not adopted, and the incumbent \model{} was retained. A continued-training run (\texttt{phoenix\_exp9\_continued\_v1}, SHA-256 \texttt{69b8c5d1...}, recorded as trained over 39,344 unique lines across 64,000 weighted draws) plateaued around 8.01\%--8.02\% validation macro CER from an 8.13\% baseline and reached 9.71\% against 9.80\% on an 800-line test. Together these suggest diminishing returns under this continued-training recipe, which motivated shifting focus to candidate selection; that suggestion is exploratory.

\subsection{Language decoding policy}
For beam candidate $y$, local language reranking uses a shallow-fusion score
\begin{equation}
S(y\mid x)=\log P_{\mathrm{CTC}}(y\mid x)+\alpha\log P_{\mathrm{LM}}(y)+\beta |y|.
\end{equation}
The precise $\alpha$, $\beta$, beam, and confidence band are decoder configuration, not learned \model{} weights. In the deployed configuration the language model is an order-8 character $n$-gram model combined with $\alpha=0.5$, $\beta=0.3$, and beam width 10, and a per-line confidence gate (default threshold 0.95) restricts automatic application to low-confidence lines; a symbol-level guard protects numbers and punctuation from destructive changes. More importantly, the LM is conditioned on domain. A general/Maghrebi model can be applied automatically only in a conservative band; an archival LM is displayed as advice; an unknown domain defaults to the visual greedy reading.

This policy follows an empirical asymmetry. On an archival development set, a domain-matched mixed LM reduced normalized CER from 11.52\% to 10.89\% and WER from 38.20\% to 31.88\%. A mismatched general LM instead raised CER to 14.78\% and WER to 44.98\%. Domain mismatch remained harmful even after visual recognition improved, so language modelling is treated as evidence, not a universal correction function.

\section{Athar: Evidence-Aware Review}
\subsection{Alternatives without false probabilities}
The interface can expose a bounded top-$k$ set (currently four in the compact review view; up to eight in analysis artifacts). Visual candidates are generated from the CTC beam without the LM. Linguistic candidates rerank those paths with a selected local LM. A relative beam score answers ``which candidate is preferred within this search?''; it does not answer ``what is the probability that this text is correct?'' The UI therefore uses rank and qualitative evidence labels rather than a percentage confidence claim.

The current text is also not duplicated as an unrelated candidate. Candidate caches are keyed by model hash, line-image hash, polygon hash, beam size, $k$, and decoder version. This fixed a practical failure in which alternatives from a different line could appear under the current transcript.

\subsection{Retrieval and attribution states}
Retrieval-augmented correction (RAC) searches a local library chosen for the manuscript project. Sources may be entered as \texttt{reference | text}, imported from an OpenITI file or approved URL, or represented by IIIF metadata and images. IIIF resources are not treated as textual witnesses unless textual annotations are present. The deployed library is deliberately described by both the number of works and the number of searchable reference units; reference positions are not called sources.

Retrieval uses normalized character $n$-gram voting to generate candidate windows, followed by local alignment and a minimum similarity rule. Queries may include the greedy line and visual alternatives. Results aggregate by source identifier and position, not by title alone. The output state is:
\begin{itemize}
  \item \textbf{unique}: evidence supports one position strongly enough to display a reference;
  \item \textbf{ambiguous}: multiple positions remain; all bounded candidates may be displayed, but no single reference is asserted;
  \item \textbf{abstain}: no sufficiently supported match is returned.
\end{itemize}
Truncation is itself evidence against uniqueness: if the search result is cut off before all competitors are observed, the best position cannot be declared unique merely because it leads the visible list.

\subsection{Researcher decisions and export}
The evidence panel separates visual, linguistic, source, and external-LLM channels. Scores from different channels are not numerically compared as if they shared a scale. A review queue first ranks lines with fast CTC-derived signals, then enriches visible items progressively and caches the evidence. Experiments found that multi-signal enrichment did not reliably outperform the CTC-only ordering; the current contribution is therefore workflow, explanation, and state management rather than an unsupported ranking claim.

A researcher decision supersedes every automatic channel. Reviewed lines can be removed from the active queue or shown again. PAGE-XML export preserves polygons and baselines; TEI export preserves the accepted text; the default training bundle includes only human-approved decisions. A source match or an unreviewed model output is never promoted to ground truth automatically, avoiding a self-training loop in which the system learns its own errors.

\section{Experimental Design}
\subsection{Metrics}
Let $r$ be the reference, $h$ the hypothesis, and $d(\cdot,\cdot)$ Levenshtein distance. We report
\begin{equation}
\cer=\frac{d_{\mathrm{char}}(r,h)}{|r|},\qquad
\wer=\frac{d_{\mathrm{word}}(r,h)}{|r|_{\mathrm{word}}}.
\end{equation}
Raw CER and WER count digits, punctuation, and the supplied transcription policy. Arabic-normalized scores are used only as diagnostics and are never compared with another system's raw score. Word accuracy and exact-line accuracy are reported where available because equal CER can conceal different correction workloads.

\subsection{Evidence tiers}
Table~\ref{tab:evidence-tiers} is a claim boundary, not merely presentation. The sealed evaluation determines the release claim. Development guards diagnose domain behavior and select checkpoints. The additional Baseer server benchmark is hypothesis-generating until line predictions, sample counts, preprocessing, and executable evaluation are archived.

\begin{table}[t]
\centering
\caption{Evidence tiers used in this paper.}
\label{tab:evidence-tiers}
\small
\begin{tabular}{p{0.25\columnwidth}p{0.67\columnwidth}}
\toprule
\textbf{Tier} & \textbf{Permitted interpretation} \\
\midrule
Sealed evaluation & Frozen exp8 vs.\ exp9; raw references; greedy decoding; no tuning after opening. Primary release evidence. \\
Development diagnostic & Identical line inputs and protocol across models; some guards influenced selection. Domain diagnosis, not final generalization. \\
Historical / server diagnostic & Aggregated diagnostic result preserved from historical development records; informative for decoding analysis, not primary held-out generalization. \\
Additional server benchmark & Owner-run Phoenix--Baseer aggregates; incomplete reproducibility bundle. Preliminary comparison only. \\
\bottomrule
\end{tabular}
\end{table}

\subsection{Datasets and split hygiene}
\begin{table*}[t]
\centering
\caption{Dataset roles in the current study. Exact training counts depend on replay sampling; the table records the unit of separation and experimental role.}
\label{tab:data}
\small
\resizebox{\textwidth}{!}{%
\begin{tabular}{p{0.12\textwidth}p{0.24\textwidth}p{0.18\textwidth}p{0.16\textwidth}p{0.20\textwidth}}
\toprule
\textbf{Dataset} & \textbf{Domain} & \textbf{Training/adaptation role} & \textbf{Separation unit} & \textbf{Evaluation role} \\
\midrule
Muharaf & 19th--20th c. archival handwriting & Replay and archival adaptation & Document & Development guard; not a new sealed result \\
RASAM & Maghrebi manuscripts & Core training and replay & Manuscript/page manifests & Development guard and historical results \\
TariMa & Diverse Maghrebi/historical hands & Replay & Manuscript where available & 164-line sealed release set \\
Agapet & 13th--17th c. Christian Arabic & SA-418 adaptation & Manuscript & Sin423 + BnF Arabe 76 sealed set \\
Omar & 1951--1965 archival memoirs & Six-document adaptation side & Document & Eleven-document sealed set \\
\bottomrule
\end{tabular}
}
\end{table*}

The official splits of several datasets are not manuscript-independent; consequently, results on those partitions may not estimate generalization to unseen manuscripts and are not directly interchangeable with our document/manuscript-held-out evaluation. For this study, Omar was regrouped at document level before adaptation: six documents on the calibration/training side and eleven held-out documents. Agapet was split by manuscript: SA-418 for adaptation and Sin423 plus BnF Arabe 76 for held-out evaluation. File hashes, identifiers, raw and normalized text overlap, and perceptual image warnings were audited. Perceptual hashes were treated as review signals, not automatic evidence of duplication.

``Sealed'' here means operationally that, before the held-out sets were opened, the two model hashes, the decoder mode (greedy, no LM), the raw-reference policy, and the membership manifests were fixed. After opening, no model or decoding parameter was tuned on the held-out results.

\section{Results}
\subsection{Sealed release evaluation}
\begin{figure}[t]
  \centering
  \includegraphics[width=\columnwidth]{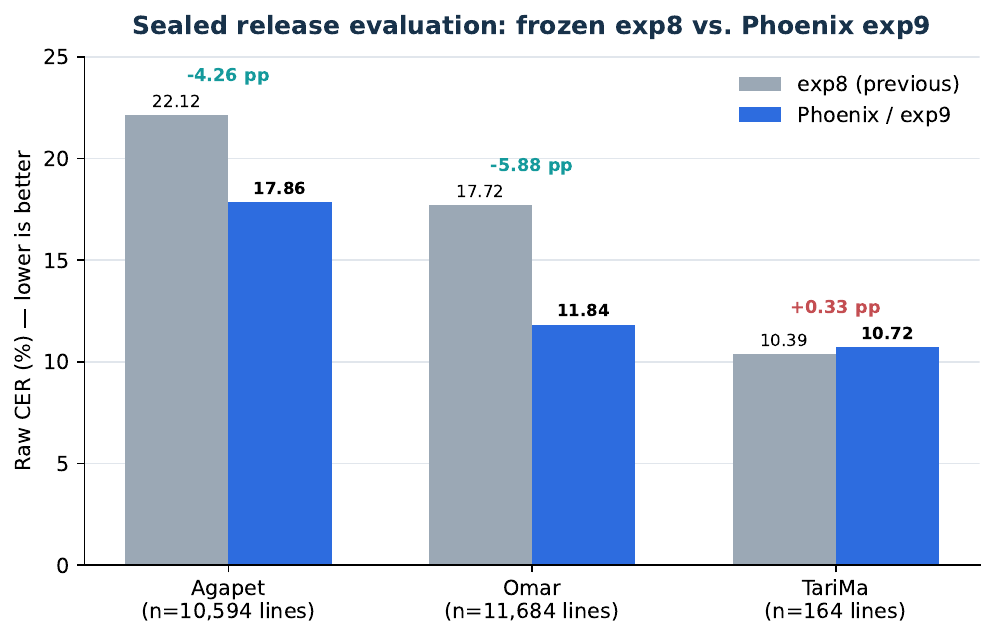}
  \caption{CER on the three sealed release sets. The TariMa regression is included in the release decision.}
  \label{fig:sealed}
\end{figure}

Table~\ref{tab:sealed} and Figure~\ref{fig:sealed} contain the strongest evidence. \model{} improved Agapet by 4.26 CER points and Omar by 5.88 points. It regressed on TariMa by 0.33 points, which is reported rather than hidden. Both Agapet manuscripts and all eleven Omar documents improved; the average therefore does not conceal a harmed held-out document within those two sets.

\begin{table*}[t]
\centering
\caption{Sealed release comparison. Raw references, greedy decoding, no language model. Lower CER/WER is better; higher word accuracy is better.}
\label{tab:sealed}
\small
\begin{tabular}{lrrrrrr}
\toprule
\textbf{Set} & \textbf{Lines} & \textbf{exp8 CER} & \textbf{exp9 CER} & \textbf{$\Delta$ CER} & \textbf{exp9 WER} & \textbf{Word acc. exp8 $\rightarrow$ exp9} \\
\midrule
Agapet & 10,594 & 22.124 & \textbf{17.861} & \textbf{$-4.263$} & 58.798 & 23.91 $\rightarrow$ \textbf{33.23} \\
Omar & 11,684 & 17.718 & \textbf{11.835} & \textbf{$-5.883$} & 42.917 & 33.40 $\rightarrow$ \textbf{46.83} \\
TariMa & 164 & \textbf{10.386} & 10.718 & $+0.332$ & 38.886 & 51.32 $\rightarrow$ 49.23 \\
\bottomrule
\end{tabular}
\end{table*}

For the two large sealed sets together (22,278 lines; 1,343,406 reference characters), character-weighted CER decreased from 19.9835\% to 14.9333\%. This is an absolute change of $-5.0502$ points and a 25.27\% relative reduction in character errors. The combined statistic excludes the small TariMa set to avoid letting a 164-line domain either disappear inside a 22k-line aggregate or receive the same weight as a 10k-line set; TariMa is instead reported explicitly as its own guardrail.

\subsection{Same-protocol development diagnostic}
\label{sec:dev}
The controlled development comparison in Table~\ref{tab:dev} uses identical pre-cropped line images, raw references, and Kraken greedy decoding for the original Muharaf recognizer, exp6, exp8, and exp9. It answers whether model differences persist when input and metric definitions are held fixed. It does not answer independent final generalization because some guards influenced exp9 selection.

\begin{table*}[t]
\centering
\caption{Same-protocol development diagnostic. Each cell is CER / word accuracy (\%). Lowest CER is bold.}
\label{tab:dev}
\small
\begin{tabular}{lrrrrr}
\toprule
\textbf{Guard} & \textbf{Lines} & \textbf{Original Muharaf} & \textbf{exp6} & \textbf{exp8} & \textbf{Phoenix / exp9} \\
\midrule
Agapet & 991 & 33.15 / 16.62 & 23.82 / 24.71 & 20.08 / 33.58 & \textbf{9.47} / 67.40 \\
Omar & 1,143 & 9.26 / 60.08 & 26.88 / 21.25 & 12.06 / 53.90 & \textbf{8.91} / 63.95 \\
RASAM & 1,789 & 39.01 / 12.06 & 9.04 / 66.78 & 7.95 / 69.90 & \textbf{7.67} / 70.46 \\
Muharaf & 920 & 13.28 / \textbf{64.15} & 36.47 / 15.74 & 12.72 / 58.30 & \textbf{12.31} / 59.32 \\
\midrule
Unweighted macro & 4,843 & 23.68 / 38.23 & 24.05 / 32.12 & 13.20 / 53.92 & \textbf{9.59} / 65.28 \\
\bottomrule
\end{tabular}
\end{table*}

\begin{figure}[t]
  \centering
  \includegraphics[width=\columnwidth]{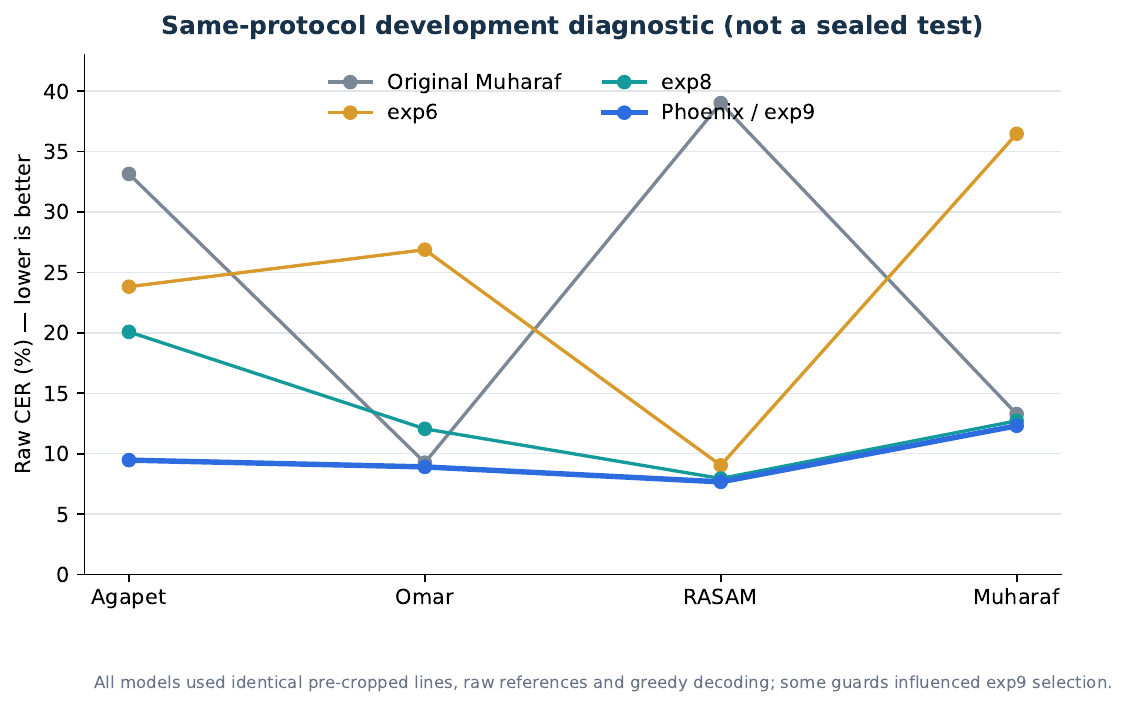}
  \caption{Controlled development CER. The figure is diagnostic, not a sealed leaderboard.}
  \label{fig:dev}
\end{figure}

\model{} has the lowest CER on four of four comparable guards and the highest word accuracy on three of four. Its unweighted macro CER is 9.59\%, compared with 23.68\% for the original Muharaf model (59.5\% relative reduction under this protocol). The Muharaf-domain row demonstrates why both character and word metrics matter: \model{} has lower CER (12.31 vs. 13.28), while the original model has higher word accuracy (64.15 vs. 59.32). A 369-line TariMa guard is excluded because its held manuscripts were used to train exp6/exp8, although they were independent for exp9; including it would not be a fair four-model comparison.

\subsection{Language model diagnostics}
Table~\ref{tab:lm} reports a consumed development set of 920 archival lines with the frozen exp8 visual model. The matched mixed LM improves CER, WER, and exact lines; the mismatched general LM is strongly harmful. The result motivates domain routing and reversibility, not an unconditional LM claim. The per-decoder values in this table are preserved from the frozen research record; their raw per-line artifact is not retained in the current repository, so the table is reported as a historical diagnostic rather than a locally re-runnable result.

\begin{table}[t]
\centering
\caption{Domain-conditioned LM diagnostic on 920 archival development lines.}
\label{tab:lm}
\small
\resizebox{\columnwidth}{!}{%
\begin{tabular}{lrrr}
\toprule
\textbf{Decoder} & \textbf{Norm. CER} & \textbf{Norm. WER} & \textbf{Exact lines} \\
\midrule
Greedy & 11.52 & 38.20 & 10.5 \\
Archive LM & 11.04 & 32.09 & 14.6 \\
Mixed LM & \textbf{10.89} & \textbf{31.88} & \textbf{14.8} \\
Mismatched general LM & 14.78 & 44.98 & 5.1 \\
\bottomrule
\end{tabular}
}
\end{table}

The mixed LM improved 371 lines and harmed 310, but the magnitudes favored improvement: 940 character errors were removed and 637 added (net $+303$ correct characters). It repaired 60 previously exact lines and broke 21 (net $+39$ exact lines). No simple inference-visible change-size or frequency gate separated the 21 harmful cases from helpful corrections. The archival LM is therefore advisory in \system{}. On the 1,789-line RASAM development guard, conservative gated LM decoding produced only negligible recognition changes, supporting the conclusion that the large held-out gains arise from the visual recognizer rather than post-hoc language correction.

\subsection{Retrieval ablation}
In a development-only RAC ablation, 192 positive lines and 162 hard negative/context lines were drawn from 25 pages. The ablation used the earlier \texttt{candidate\_rasam609\_9f30ebaf} recognizer (SHA-256 prefix \texttt{9f30ebafdf044d22}) to generate visual alternatives; it therefore evaluates the retrieval strategy rather than end-to-end performance of the final \model{} checkpoint. The single greedy query retrieved correct text for 69.8\% of positive lines at 100\% text precision. Querying the top five visual candidates raised recall to 80.7\%, recovering 21 lines and breaking none. Of 162 negatives, 161 abstained and one returned an ambiguous repeated formula; none produced a unique attribution requiring review. These figures test retrieval against transcriptions aligned to the same manuscripts, not against a different critical edition. Orthographic and editorial variation between witnesses remains an open problem.

\subsection{Preliminary comparison with a large VLM}
Table~\ref{tab:baseer} records an additional owner-run server comparison with Baseer\_\_Nakba, a model derived from the Baseer vision-language family~\cite{misraj2025baseer}. The supplied aggregates lack raw line predictions, sample counts, a preprocessing manifest, and an executable evaluation bundle. They are included because they motivate the efficiency question, but they are not sealed or independently reproduced and must not be mixed with Tables~\ref{tab:sealed} or~\ref{tab:dev}. \texttt{Baseer\_\_Nakba} is adapted to the NAKBA Arabic manuscript task, and the Omar Al-Saleh collection is that task's shared-task corpus~\cite{hamoud2026nakba}; Omar is therefore a task-matched domain for that model rather than an unfamiliar one. Whether its training data overlaps the specific Omar subset used in this benchmark cannot be reconstructed from the retained server artifacts, and we make no claim of contamination. We report Omar separately for this reason and, alongside the four-domain macro, give a descriptive macro over the three remaining domains. Neither aggregate is treated as primary comparative evidence.

\begin{table*}[t]
\centering
\caption{Preliminary owner-run server comparison; exploratory evidence tier. Raw predictions, sample counts, the preprocessing manifest, and the WER and exact-line denominators were not retained, so these aggregates are not independently reproducible. The four-domain macro CER is an unweighted mean across Agapet, Muharaf, Omar, and RASAM and includes Omar, which is task-matched to \texttt{Baseer\_\_Nakba} through the NAKBA shared task~\cite{hamoud2026nakba}. A descriptive macro over the three remaining domains is reported alongside it. Baseer attains the lower Omar CER and the higher exact-line accuracy. These values must not be combined with Table~\ref{tab:sealed} or Table~\ref{tab:dev}.}
\label{tab:baseer}
\small
\begin{tabular}{lrrr}
\toprule
\textbf{Metric} & \textbf{Phoenix} & \textbf{Baseer\_\_Nakba} & \textbf{Observed difference} \\
\midrule
Macro CER, 4 domains (incl.\ task-matched Omar) & \textbf{9.71\%} & 28.67\% & $-18.96$ points \\
Macro CER, 3 domains (non-Omar, descriptive) & \textbf{10.65\%} & 38.07\% & $-27.42$ points \\
WER (aggregation denominator not supplied) & \textbf{34.37\%} & 59.52\% & $-25.15$ points \\
Exact-line accuracy & 13.38\% & \textbf{26.38\%} & Baseer $+13.00$ points \\
Parameters & \textbf{4.99M} & $\approx$3.75B & Phoenix $\approx752\times$ smaller \\
Model file & \textbf{19.94 MB} & $\approx$7.53 GB & Phoenix $\approx378\times$ smaller \\
\bottomrule
\end{tabular}
\end{table*}

Phoenix had lower CER on Agapet (11.13 vs. 35.65), Muharaf (11.93 vs. 25.51), and RASAM (8.90 vs. 53.06). Baseer had much lower CER on Omar (0.48 vs. 6.88) and higher overall exact-line accuracy. The defensible interpretation is conditional: a compact specialist can be competitive or superior under several domains and far cheaper to store, while a large VLM can dominate a task-matched domain and produce more exact lines. Archiving the missing evaluation artifacts is required before this comparison becomes publication-grade evidence.

\subsection{Candidate-selection headroom and oracle analysis}
To investigate where remaining transcription errors originate, a balanced five-domain candidate pool was assembled across Agapet, Muharaf, Omar, RASAM, and TariMa (1,500 line groups, 300 per source). After excluding 200 previously tuned lines, the set was partitioned into 1,250 training and 250 isolated development lines (50 per source, with no PAGE-XML overlap between splits). The exact producer checkpoint for this historical diagnostic pool could not be recovered from the retained logs. We therefore treat these results strictly as a development diagnostic.

On the frozen 250-line development set, greedy decoding achieves 8.131\% CER and standard beam decoding (beam width 10) reaches 7.966\% CER. When selecting the candidate nearest to the ground truth from the top 25 beam paths, the diagnostic upper bound (\textbf{Oracle@25}) drops to \textbf{5.820\% CER}. This exposes a substantial recoverable headroom of \textbf{2.146 CER points} (a 26.9\% relative error reduction).

Crucially, Oracle@25 is not a deployable decoder; it uses ground truth solely to diagnose candidate availability. The oracle removes 26.9\% of the character errors remaining under beam decoding, showing that a substantial fraction of residual character error is recoverable from existing $N$-best hypotheses.

\begin{figure}[t]
  \centering
  \includegraphics[width=\columnwidth]{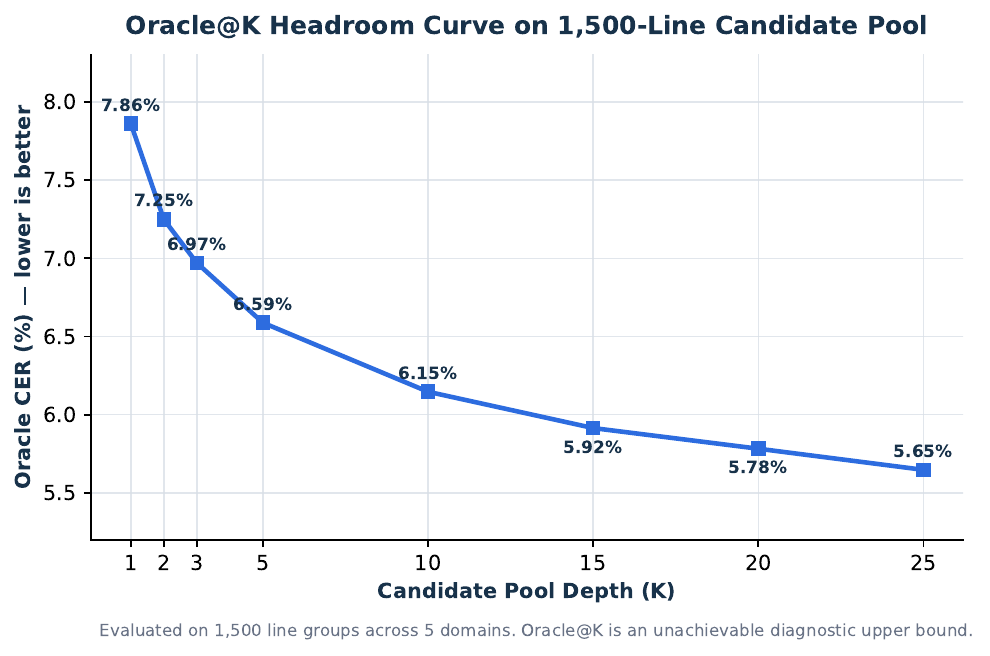}
  \caption{Oracle@K headroom curve evaluated on the full 1,500-line candidate pool across five domains. Candidate headroom expands with candidate depth but shows diminishing returns with increasing $K$.}
  \label{fig:oracle_k}
\end{figure}

Figure~\ref{fig:oracle_k} plots the Oracle@K curve as candidate depth $K$ increases from 1 to 25 on the full 1,500-line candidate pool ($K=1$: 7.860\%, $K=5$: 6.588\%, $K=15$: 5.915\%, $K=25$: \textbf{5.648\%}). The curve reveals steep initial gains from $K=1$ to $K=5$, followed by diminishing returns beyond $K=15$. (Note: the full 1,500-pool Oracle@25 value of 5.648\% is computed over the entire pool, whereas the isolated 250-dev Oracle@25 benchmark is 5.820\%). Per-source breakdown confirms that headroom exists across all five domains: Agapet (2.700\pp{} gap), Muharaf (2.328\pp{}), Omar (2.578\pp{}), RASAM (1.969\pp{}), and TariMa (1.486\pp{}). The normalization diagnostic reduces the oracle gap by only 14.64\% ($2.195\pp{} \rightarrow 1.873\pp{}$), leaving most of the recoverable gap unexplained by the normalized orthographic conventions considered here.

\subsection{Consolidated decoder and quality estimation benchmark}
We systematically evaluated whether auxiliary sequence rerankers, candidate consensus, or visual quality estimators could bridge this 2.15\pp{} oracle gap. Table~\ref{tab:decoders} presents the comparative evaluation on the frozen 250-line development diagnostic.

\begin{table}[t]
\centering
\caption{Consolidated decoder, neural reranker, and quality estimator benchmark on the frozen 250-line development diagnostic (the exact producer checkpoint could not be recovered from the retained historical logs; Oracle@25 is an unachievable diagnostic upper bound).}
\label{tab:decoders}
\small
\resizebox{\columnwidth}{!}{%
\begin{tabular}{lrp{0.48\columnwidth}}
\toprule
\textbf{Method} & \textbf{Dev CER (\%)} & \textbf{Scientific Interpretation} \\
\midrule
Greedy & 8.131 & Baseline local path selection \\
Beam (width 10) & 7.966 & Standard CTC beam search \\
Length-only control & 7.977 & Length normalization alone does not explain gains \\
Neural Reranker v1 & \textbf{7.883} & BiGRU with listwise soft-target training (small dev gain, $-0.083\pp{}$) \\
Neural Reranker v2 & 7.901 & Hybrid listwise + pairwise loss (does not outperform v1) \\
Medoid consensus & 8.007 & Hypothesis-space consensus harms top-1 selection \\
MBR ($T=5$) & 8.009 & Minimum Bayes risk consensus does not identify ground truth \\
E1 pure & 8.727 & Forced-aligned CTC posterior features alone \\
E1 + CTC fusion & 7.965 & Posterior QE practically ties baseline beam ($-0.001\pp{}$) \\
E2a pure & 8.763 & Pre-CTC hidden-state pooling ($H \in \mathbb{R}^{T \times 400}$) \\
E2a + CTC fusion & 7.967 & Local hidden-state QE practically ties baseline beam ($+0.001\pp{}$) \\
\midrule
\textbf{Oracle@25} & \textbf{5.820} & \textbf{Diagnostic upper bound (unachievable)} \\
\bottomrule
\end{tabular}
}
\end{table}

\textbf{Neural Text Rerankers (v1 and v2):} A bidirectional GRU reranker was trained with listwise soft targets on 1,250 training lines using normalized CTC score, normalized LM score, CTC gap, relative length, edit distance to greedy, beam rank, and candidate consensus features. Reranker v1 achieved 7.883\% CER on dev (a small development improvement of $-0.083\pp{}$ over beam). A hybrid listwise/pairwise variant (v2) reached 7.901\%, failing to improve over v1. A length-only control reached 7.977\%, suggesting that v1's minor gain is not purely length normalization.

\textbf{Candidate Consensus (MBR and Medoid):} Minimum Bayes Risk decoding ($T=5$) and Medoid consensus yielded 8.009\% and 8.007\% CER respectively, worsening accuracy relative to standard beam (7.966\%). In manuscript HTR, majority consensus among beam paths often reinforces correlated visual/CTC ambiguities rather than isolating the correct reading.

\textbf{Visual Quality Estimation (E1 and E2):} We tested two explicit quality estimation formulations:
\begin{itemize}
  \item \textbf{E1 (CTC Posterior QE):} Extracted frame-level character duration, entropy, label margins, and blank probabilities from forced alignment. While alignment succeeded across all lines (0 failures), E1+CTC fusion achieved 7.965\% CER, practically tying beam search (7.966\%).
  \item \textbf{E2 (Pre-CTC Hidden-State Visual QE):} Extracted the pre-projection hidden sequence $H \in \mathbb{R}^{T \times 400}$ (the 400-dimensional BiLSTM output prior to the linear classification head) and pooled representations over character frames. On frozen dev, E2a+CTC fusion achieved 7.967\% CER.
\end{itemize}
The scientific interpretation is nuanced: these results do not imply that Phoenix hidden representations lack useful information; rather, the tested CTC-posterior and locally pooled pre-CTC hidden-state QE formulations did not materially improve candidate selection.

\textbf{Neural Character LM Rescoring:} In a separate diagnostic, a 6-layer Transformer character LM (4.891M parameters, trained on 38,872 lines) reached a strong validation perplexity of 13.04 (val NLL 2.5681). However, when used for beam shallow fusion on its held-out decoder split, it achieved 8.348\% CER compared with 8.528\% beam and 8.244\% length-only (against an Oracle@25 upper bound of 6.025\%). This demonstrates that strong intrinsic language model performance (low perplexity) does not necessarily translate into effective sequence-level HTR candidate reranking.

\section{Discussion}
\subsection{What the experiments establish}
The experimental progression establishes four major empirical findings:

First, \textbf{replay-aware adaptation succeeds for compact multi-domain HTR}. Phoenix (4.99M parameters) achieves robust accuracy across Christian Arabic, Maghrebi manuscripts, and modern archival memoirs without requiring separate domain specialists. The specialist ablation demonstrates that domain-specific routing yields only marginal gains ($0.07\text{--}0.44\pp{}$) that do not justify multi-model deployment overhead. The small TariMa regression ($+0.33\pp{}$) reinforces that multi-domain selection must balance a vector of domain guards rather than maximizing a single average.

Second, \textbf{post-release training appeared to reach diminishing returns under the tested continued-training recipe}. The extra epoch and the continued-training sweep (\texttt{continued\_v1}) plateaued around 8.02\% validation macro CER. These runs are exploratory: their executable checkpoints and full evaluation artifacts were not retained (Section~\ref{sec:replay}), so the observation rests on summary records and is not independently verifiable.

Third, \textbf{candidate selection constitutes a substantial remaining bottleneck alongside candidate generation}. While the recognizer's beam generates lower-error hypotheses (yielding a 2.15\pp{} oracle gap on dev), candidate generation itself remains an error source (5.82\% Oracle@25 CER), and automated sequence selection (rerankers, MBR, CTC-QE, hidden-state QE) recovers less than 4\% of the oracle headroom.

Fourth, \textbf{these empirical limitations directly motivate Athar's evidence-preserving review design}. Because automated sequence rerankers cannot reliably pick the best hypothesis, the transcription system must not silently overwrite text. Athar exposes bounded alternatives, preserves the raw visual reading, and maintains auditable human state.

\subsection{Research and system contributions}
The contribution is the unified synthesis of:
\begin{enumerate}
  \item replay-aware multi-domain HTR with explicit forgetting guards;
  \item document-aware and sealed evaluation claim boundaries;
  \item empirical identification of the candidate-selection oracle gap;
  \item demonstration that local text and visual quality estimators fail to bridge this gap;
  \item retrieval that separates text correction from source attribution;
  \item explicit unique/ambiguous/abstain states;
  \item auditable human decisions and human-only default training export; and
  \item provenance hashes binding model, image, geometry, decoder, and source corpus.
\end{enumerate}

\subsection{Limitations}
First, \textbf{full-page accuracy depends heavily on layout segmentation}. In a historical 13-page held-out full-pipeline probe, ordinary pages achieved $\approx 7.63\%$ CER, while two complex 1926-layout pages reached $\approx 27.30\%$ CER (overall $\approx 10.65\%$). On a difficult framed page (\texttt{logic\_16}), switching from automatic line detection (32.9\% CER) to expert manual/GT line crops reduced error to 3.2\%--3.6\% CER, demonstrating that layout failure can dominate recognition error by up to $10\times$.

Second, Agapet CER remains 17.86\%; the model is an assistant, not an automatic edition generator. Third, the small TariMa sealed set gives a weak estimate of domain variance, and the official partitions of several datasets are not manuscript-independent (potentially limiting generalization estimates on unseen manuscripts). Fourth, no uncertainty intervals are reported for the sealed comparison. The per-line prediction records for that evaluation are retained and reproduce the reported aggregates exactly, so such an analysis is possible in principle, but it requires a sampling unit and cluster structure specified in advance and is left to future work; the records are not yet publicly archived. No equivalent per-line records exist for the Baseer comparison, so no uncertainty analysis is possible there at all. Fifth, candidate scores and CTC confidence are not globally calibrated. Sixth, the source library is limited and a match to a repeated phrase cannot establish unique provenance. Seventh, the Baseer comparison is preliminary. Eighth, the model is released under a conservative non-commercial ShareAlike licence; because the exact snapshot of one training corpus could not be reconstructed, downstream users should verify upstream terms for their own copies rather than relying on the summary above. Ninth, the 250-line decoder development set was used during model tuning and comparison among reranking methods; observed differences (such as $7.966\% \rightarrow 7.883\%$) are therefore descriptive development-only findings rather than independent confirmation on an unseen held-out set. Finally, no formal researcher usability study has been conducted; \system{} is evaluated here as a technical evidence-management workflow, and usability studies remain future work.

\subsection{Future work}
The next research priorities are:
\begin{itemize}
  \item develop sequence-level cross-modal vision-language rerankers to bridge the 2.15\pp{} oracle headroom;
  \item leverage pre-CTC sequence representations (groundwork cached for 39,344 training lines in float16, 4.77 GiB) for non-local encoder re-scoring;
  \item release line-level prediction artifacts and executable manifests for manuscript-level bootstrap confidence intervals;
  \item calibrate CTC confidence by domain and study risk--coverage curves;
  \item improve automatic layout analysis on complex framed layouts;
  \item scale the local source library with rights and provenance metadata;
  \item conduct formal scholarly usability and time-to-correction studies.
\end{itemize}

\section{Reproducibility, Ethics, and Release}
The released model is \model{} (checkpoint \texttt{exp9}) with SHA-256 \nolinkurl{2896fef9d9665cbb82fba8faa3bf0c628ac6a5cc62eb678f40c707db833aebea}. The model card records Python 3.10.11, PyTorch 2.4.1+cu121, the greedy evaluation decoder, architecture metadata, data roles, licences, and claim boundaries~\cite{phoenix2026model}. Phoenix is released under CC BY-NC-SA 2.0 as a conservative project-level licensing choice. Source-dataset licensing is reported separately from the model-release decision. RASAM and TariMa are currently published under Apache-2.0 in their upstream dataset repositories; Agapet is distributed under CC BY 4.0; and the Omar Al-Saleh data are distributed under CC BY 4.0 with gated access. For Muharaf, retained project metadata records a CC BY-NC-SA family designation without a version. Because the exact Muharaf snapshot used in training could not be reconstructed, we do not assign a licence version to that training copy here.

Decoder, language-model, and retrieval configuration. The deployed language decoder is a shallow-fusion beam decoder over an order-8 character $n$-gram model with $\alpha=0.5$, $\beta=0.3$, beam width 10, and a per-line confidence gate at threshold 0.95; numbers and punctuation are protected by a symbol-level guard, and domain routing restricts automatic LM application to a conservative band (archival LM advisory, unknown domain greedy). The retrieval layer indexes normalized character 12-grams (query step 4), aligns candidates with \texttt{difflib.SequenceMatcher}, and applies a minimum similarity of 0.78, a minimum line length of 14, a maximum length change of 0.20, and a uniqueness margin of 0.02; at most eight candidate positions are displayed and truncation is reported as a lower bound. Arabic-normalized metrics, used only for diagnostics, apply Unicode NFC, fold visually equivalent letters, and drop diacritics, tatweel, Latin, digits, and punctuation; raw CER/WER remain the primary metrics and normalized scores are never compared with another system's raw score.

Availability. The Phoenix model, model card, benchmark summaries, and evaluation protocols are public at \href{https://huggingface.co/factlogic/phoenix-arabic-manuscript-htr}{huggingface.co/factlogic/phoenix-arabic-manuscript-htr}. The evaluation scripts (\texttt{scripts/cer.py}, \texttt{backend/app/services/arabic\_metrics.py}), decoder/LM configuration, and retrieval configuration are in the project repository. The per-line predictions for the held-out release evaluation are retained and reproduce every value in Table~\ref{tab:sealed}; they are not yet publicly archived and remain a pending release artifact. The per-line predictions for the Baseer server comparison were not retained and cannot be published. The two are listed separately because one is an archiving task and the other an unrecoverable gap.

\section{Conclusion}
\model{} demonstrates that a compact CNN--BiLSTM--CTC recognizer can be extended across diverse Arabic manuscript domains with replay and guard-based selection. On 22,278 large-set sealed lines it reduced character-weighted CER from 19.98\% to 14.93\% relative to the preceding checkpoint, while a separate small TariMa set recorded a 0.33-point regression. Systematic $N$-best candidate analysis reveals a 2.15\pp{} oracle headroom that automated text and quality estimation rerankers fail to bridge. \system{} addresses this candidate-selection bottleneck by turning HTR from a black-box text generator into an auditable, evidence-preserving review workflow.

\section*{Acknowledgements}
We thank the creators and annotators of Muharaf, RASAM, TariMa, Agapet, and the Omar Al-Saleh manuscript resources, and the developers of Kraken, eScriptorium, PAGE-XML, OpenITI, and IIIF.

\appendix

\section{Historical Early LM and Safety Gating Probes}
Table~\ref{tab:app_historical_lm} summarizes early language modeling benchmarks with the exp6 visual recognizer on RASAM and TariMa test splits, demonstrating early gains from character $n$-gram context ($\alpha=0.25, \text{beam}=12$).

\begin{table}[h]
\centering
\caption{Historical exp6 character $n$-gram LM benchmark.}
\label{tab:app_historical_lm}
\small
\resizebox{\columnwidth}{!}{%
\begin{tabular}{lrr}
\toprule
\textbf{Configuration} & \textbf{RASAM-test CER (\%)} & \textbf{TariMa-test CER (\%)} \\
\midrule
Greedy (no LM) & 7.48 & 10.27 \\
Order-8 LM & \textbf{6.76} & \textbf{8.68} \\
$\Delta$ & $-0.72\pp{}$ & $-1.59\pp{}$ \\
\bottomrule
\end{tabular}
}
\end{table}

Table~\ref{tab:app_gated_safety} reports the confidence-gated safety probe across clean (D1), demo (D2), and auto-segmented (D3) lines. While always-on LM damages already-accurate visual text (raising D1 CER from 0.39\% to 0.56\%), confidence gating (<0.95 threshold) prevents degradation on clean lines (0.30\%) while capturing gains on difficult lines.

\begin{table}[h]
\centering
\caption{Confidence-gating safety probe across diagnostic sets.}
\label{tab:app_gated_safety}
\small
\resizebox{\columnwidth}{!}{%
\begin{tabular}{lrrrr}
\toprule
\textbf{Decoder Mode} & \textbf{D1 Clean} & \textbf{D2 Demo} & \textbf{D3 Auto} & \textbf{Mean CER (\%)} \\
\midrule
Greedy (no LM) & 0.39 & 4.34 & 3.61 & 2.78 \\
Always-LM & 0.56 & \textbf{2.56} & \textbf{2.28} & \textbf{1.80} \\
Gated LM (conf $<0.95$) & \textbf{0.30} & 3.29 & 3.41 & 2.33 \\
\bottomrule
\end{tabular}
}
\end{table}

In a self-trained domain modeling probe, 60 non-GT pages were transcribed, yielding 418 high-confidence pseudo-labeled lines. On a logic-manuscript ground truth text, general LM perplexity (53.14) fell to 40.70 with the self-trained model, and to \textbf{20.25} with a mixed general+domain model (a 61.9\% relative perplexity reduction).

\section{Oracle Headroom Error Concentration and Source Breakdown}
Table~\ref{tab:app_oracle_concentration} records the cumulative concentration of recoverable oracle errors across percentiles on the 250 development lines. The distribution confirms that recoverable errors are broadly distributed rather than concentrated in a few outlier lines.

\begin{table}[h]
\centering
\caption{Cumulative concentration of recoverable oracle errors on the 250-line development set.}
\label{tab:app_oracle_concentration}
\small
\resizebox{\columnwidth}{!}{%
\begin{tabular}{lr}
\toprule
\textbf{Top Percentile of Lines} & \textbf{Share of Recoverable Errors (\%)} \\
\midrule
Top 5\% & 12.09 \\
Top 10\% & 21.63 \\
Top 20\% & 37.61 \\
Top 30\% & 53.60 \\
Top 50\% & 76.24 \\
\bottomrule
\end{tabular}
}
\end{table}

Table~\ref{tab:app_oracle_sources} breaks down baseline beam CER against Oracle@25 across the five represented manuscript domains on the full 1,500-line candidate pool (300 lines per domain).

\begin{table}[h]
\centering
\caption{Oracle@25 headroom broken down by manuscript domain across the full 1,500-line candidate pool (300 lines per domain).}
\label{tab:app_oracle_sources}
\small
\resizebox{\columnwidth}{!}{%
\begin{tabular}{lrrr}
\toprule
\textbf{Domain} & \textbf{Beam CER (\%)} & \textbf{Oracle@25 CER (\%)} & \textbf{Gap ($\Delta$ pp)} \\
\midrule
Agapet & 8.923 & 6.223 & 2.700 \\
Muharaf & 12.232 & 9.904 & 2.328 \\
Omar & 7.858 & 5.280 & 2.578 \\
RASAM & 7.424 & 5.455 & 1.969 \\
TariMa & 2.862 & 1.376 & 1.486 \\
\midrule
Unweighted Mean & 7.860 & 5.648 & 2.212 \\
\bottomrule
\end{tabular}
}
\end{table}

\section{Neural Character Language Model Specifications}
The neural character LM is a 6-layer autoregressive Transformer with 8 attention heads, model dimension $d=256$, feed-forward dimension 1024, dropout 0.15, and 4,891,089 parameters. Training data comprised 38,872 lines and 4,212 validation lines across the five manuscript collections (after purging 472 exact duplicates). Vocabulary size is 81 tokens (77 unique character glyphs). At best epoch 12, validation loss reached NLL 2.5681 and perplexity 13.04. Table~\ref{tab:app_neural_ppl} lists per-domain validation perplexities.

\begin{table}[h]
\centering
\caption{Per-domain validation perplexities for the neural character LM.}
\label{tab:app_neural_ppl}
\small
\resizebox{\columnwidth}{!}{%
\begin{tabular}{lrr}
\toprule
\textbf{Domain} & \textbf{Validation NLL} & \textbf{Validation PPL} \\
\midrule
Agapet & 2.472 & 11.85 \\
Muharaf & 2.640 & 14.02 \\
Omar & 2.700 & 14.87 \\
RASAM & 2.520 & 12.43 \\
TariMa & 2.506 & 12.26 \\
\midrule
Overall & \textbf{2.568} & \textbf{13.04} \\
\bottomrule
\end{tabular}
}
\end{table}

\section{Character Error Anatomy by Manuscript Domain}
Analysis of character substitution distributions reveals domain-specific challenges:
\begin{itemize}
  \item \textbf{Agapet (Christian Arabic):} Dot-confusable character errors account for $\approx 41.88\%$ of all errors, reflecting sparse or non-standard diacritical pointing.
  \item \textbf{Omar (Archival Memoirs):} Hamza/alif orthographic alternations account for $\approx 26.21\%$ of substitutions.
  \item \textbf{RASAM (Maghrebi Manuscripts):} General character substitutions account for $\approx 30.62\%$, Maghrebi dot placement confusions (e.g., $f\bar{a}'$ with one sublinear dot vs.\ $q\bar{a}f$ with one supralinear dot) account for $\approx 20.60\%$, and word-boundary spacing errors account for $\approx 18.97\%$.
  \item \textbf{TariMa (Maghrebi/Historical):} Dot confusions account for $\approx 27.34\%$ and other substitutions account for $\approx 26.17\%$.
  \item \textbf{Common Confusion Pairs:} Across all collections, dominant confusions include $l\bar{a}m \rightarrow k\bar{a}f$, $d\bar{a}l \rightarrow r\bar{a}'$, $r\bar{a}' \rightarrow n\bar{u}n$, $r\bar{a}' \rightarrow d\bar{a}l$, and $n\bar{u}n \rightarrow l\bar{a}m$.
\end{itemize}

\section{Implementation Groundwork: Pre-CTC Hidden-State Cache}
As implementation groundwork for future sequence-level visual-linguistic rerankers, the pre-projection hidden sequence $H \in \mathbb{R}^{T \times 400}$ (4-layer BiLSTM output, 400 dimensions per frame) was extracted and cached in float16 for all 39,344 training lines (occupying 4.77 GiB on disk). The downstream $F1$ sequence-level cross-modal reranker was not trained during this project phase, and no experimental gains are claimed for this groundwork.

\balance
\bibliographystyle{unsrtnat}
\bibliography{references}

\end{document}